\documentclass[sn-mathphys-ay]{sn-jnl}

\usepackage{graphicx}%
\usepackage{multirow}%
\usepackage{amsmath,amssymb,amsfonts}%
\usepackage{amsthm}%
\usepackage{mathrsfs}%
\usepackage[title]{appendix}%
\usepackage{xcolor}%
\usepackage{textcomp}%
\usepackage{manyfoot}%
\usepackage{booktabs}%
\usepackage{algorithm}%
\usepackage{algorithmicx}%
\usepackage{algpseudocode}%
\usepackage{listings}%
\usepackage{graphicx}  
\usepackage{float}  
\usepackage{subfigure}  
\usepackage{subcaption}  
\usepackage{placeins}
\usepackage{array} 
\usepackage{hyperref}

\theoremstyle{thmstyleone}%
\theoremstyle{thmstyletwo}%

\theoremstyle{thmstylethree}%

\begin{document}

\title[Enhanced PEC-YOLO for Detecting Improper Safety Gear Wearing Among Power Line Workers]{Enhanced PEC-YOLO for Detecting Improper Safety Gear Wearing Among Power Line Workers}

\author[1,2,3]{\fnm{Chen} \sur{Zuguo}}\email{zg.chen@hnust.edu.cn}

\author*[3]{\fnm{Kuang} \sur{Aowei}}\email{793526042@qq.com}

\author[1,3]{\fnm{Huang} \sur{Yi}}

\author[1,3]{\fnm{Jin} \sur{Jie}}

\affil[1]{\orgdiv{Sanya Research Institute}, \orgname{Hunan University of Science and Technology}, \orgaddress{ \city{Sanya}, \postcode{572024}, \country{China}}}

\affil[2]{\orgdiv{Shenzhen Institutes of Advanced Technology}, \orgname{Chinese Academy of Sciences}, \orgaddress{ \city{Xiangtan}, \postcode{518055}, \country{China}}}

\affil*[3]{\orgdiv{School of Information and Electrical Engineering}, \orgname{Hunan University of Science and Technology}, \orgaddress{\city{Xiangtan}, \postcode{411201}, \country{China}}}


\abstract{To address the high risks associated with improper use of safety gear in complex power line environments, where target occlusion and large variance are prevalent, this paper proposes an enhanced PEC-YOLO object detection algorithm. The method integrates deep perception with multi-scale feature fusion, utilizing PConv and EMA attention mechanisms to enhance feature extraction efficiency and minimize model complexity. The CPCA attention mechanism is incorporated into the SPPF module, improving the model's ability to focus on critical information and enhance detection accuracy, particularly in challenging conditions. Furthermore, the introduction of the BiFPN neck architecture optimizes the utilization of low-level and high-level features, enhancing feature representation through adaptive fusion and context-aware mechanism. Experimental results demonstrate that the proposed PEC-YOLO achieves a 2.7\% improvement in detection accuracy compared to YOLOv8s, while reducing model parameters by 42.58\%. Under identical conditions, PEC-YOLO outperforms other models in detection speed, meeting the stringent accuracy requirements for safety gear detection in construction sites. This study contributes to the development of efficient and accurate intelligent monitoring systems for ensuring worker safety in hazardous environments. Code is available at \href{https://github.com/kuang-aowei/PEC-YOLO}{https://github.com/kuang-aowei/PEC-YOLO}}

\keywords{Lightweight Algorithm, Object Detection, YOLO, PConv, Attention Mechanism, BiFPN}



\maketitle

\section{Introduction}\label{sec1}

The safety of power production is crucial for the stable operation of power systems, and ensuring the safety of construction workers is an urgent need at present. The power construction environment is complex and high-risk, with safety hazards emerging continuously\citep{bib1}. Traditional manual inspection-based supervision methods are inefficient, costly, and prone to high error rates. Achieving effective and precise intelligent monitoring is made possible through the utilization of image recognition\citep{bib2} and object detection techniques\citep{bib3}. These methods do not require direct contact with personnel and can effectively prevent safety accidents. This method not only reduces labor costs and increases efficiency for enterprises, but also requires no modification of existing monitoring systems, offering broad research and application prospects.

Object detection methods can be classified into two primary types according to different technical strategies: traditional methods rooted in machine learning and advanced deep learning methodologies. Traditional machine learning techniques predominantly depend on pre-designed features and standard classifiers like SIFT(Scale-Invariant Feature Transform), SURF(Speeded Up Robust Features), and HOG(Histogram of Oriented Gradients). Ke Y et al. improved the local image descriptors used in SIFT by encoding salient aspects of image gradients in the neighborhood of feature points\citep{bib4}. Fan J et al. introduced a random sample consensus (RANSAC) object retrieval matching strategy using SURF for object tracking\citep{bib5}. Lin Xiaolin et al. utilized Harris detection and Histogram of Oriented Gradients (HOG) features to describe images, and then employed Support Vector Machines (SVM) for object detection\citep{bib6}. Pan utilized Gaussian Mixture Models to detect moving objects and perform Canny edge detection. He also implemented an eight-neighbor algorithm to extract various details served as auxiliary means for precise recognition, enabling the inspection of safety belt status of power site workers\citep{bib7}. Although these methods are computationally simple and highly interpretable, the need to manually adjust features to suit different scenarios results in weak generalization capabilities and insufficient adaptability to complex environments. Additionally, reliance on domain experts for feature adjustment results in high labor costs. 
 
Methods that utilize deep learning harness deep neural networks to autonomously learn and derive features from images, thereby removing the necessity for manually crafted features. These deep learning approaches can generally be divided into two main categories: two-stage algorithms and one-stage algorithms. Two-stage algorithms initially produce a set of potential regions before proceeding with classification and accurate bounding box adjustment on these regions. The R-CNN family, comprising R-CNN, Faster R-CNN, and Cascade R-CNN, exemplifies this methodology. An oriented region proposal network was introduced to optimize the R-CNN object detection model, achieving high accuracy and efficiency in two-stage oriented object detection\citep{bib8}. By excluding extraneous background areas, the algorithm's detection accuracy was significantly enhanced, leading to superior performance in identifying small objects within complex environments\citep{bib9}. The Faster R-CNN algorithm was further improved by incorporating Soft-NMS (Soft Non-Maximum Suppression) and OHEM (Online Hard Example Mining) modules, resulting in better object detection accuracy\citep{bib10}. A refined Cascade R-CNN algorithm achieved multi-level feature fusion through an improved recursive feature pyramid module, resulting in better performance in handling overlapping targets\citep{bib11}. However, the large model parameters and slower detection speed of two-stage object detection algorithms limit their deployment efficiency in real-time detection at power sites. This makes them less suitable for the urgent need for real-time and flexible deployment in modern power site safety equipment detection.

Unlike two-stage methods, one-stage algorithms eliminate the step of producing region proposals beforehand. By concatenating multi-scale features from different layers as context, CE-SSD enhanced the focus on small objects and strengthened contextual information. However, this method had a suboptimal multi-object detection performance in complex power site environments\citep{bib12}. The Efficient Attention Pyramid Transformer (EAPT) introduced Deformable Attention and an Encode-Decode Communication Module to improve vision transformer performance. However, EAPT focused on architectural innovation, lacking considerations for real-world scenarios like detecting violations in complex power line operation backgrounds.\citep{bib13}. BaGFN introduced an Attentive Graph Fusion module and a Broad Cross module to enhance high-order feature interaction modeling, significantly improving feature representation accuracy and flexibility. However, its optimization focuses primarily on feature engineering and lacks specific considerations for real-world applications, such as handling complex backgrounds and multi-scale object detection.\citep{bib14}. By using a deep feature extraction module 	and a mutual attention token selection module (MATS), the Transformer architecture enhanced the detection and localization capabilities. However, this method did not involve obtaining spatio-temporal information features and thus cannot be directly applied to the field of video anomaly monitoring.\citep{bib15}. ResNet50 was proposed as the feature extraction backbone network for YOLOv5, achieving high-precision detection of safety helmets, safety clothing, and poles in power site operations. However, this method failed to effectively analyze and address object occlusion issues in complex environments\citep{bib16}. MobileNetv3 served as the foundation for feature extraction in YOLOv5s, with the intention of trimming down the model parameters and size, consequently bolstering detection speed. However, in the highly variable environments of power construction sites, complex scenarios still presented instances of missed detection\citep{bib17}. YOLOv5 algorithm was enhanced by incorporating an asymmetric convolution module to improve the discriminability of target features. The algorithm didn?t address issues such as large target variance and target occlusion in specific scenarios\citep{bib18}. An algorithm incorporated an attention mechanism based on channel reorganization. Furthermore, it incorporated a Res-PANet design to amplify feature extraction and multi-scale feature integration. However the study did not explore the possibility of its real-time application and edge deployment\citep{bib19}. The DFP-YOLO algorithm reduced model parameters and complexity using C3\_D and C3\_F modules. However, the model's focus on reducing size and complexity may compromise its robustness in highly varied or cluttered industrial settings.\citep{bib20}.

Although there has been some research on equipment detection in power construction sites, the aforementioned methods often have limitations. These limitations become evident when faced with complex backgrounds, target occlusion, and large target variances in these environments. In order to tackle the issues mentioned above, the PEC-YOLO model is proposed to detect instances of improper protective gear usage among electrical workers in this paper. The model employs PConv and EMA attention mechanisms to enhance the performance of the feature extraction network. By deploying pointwise convolution and efficient feature aggregation techniques, the model effectively minimizes the number of parameters and computational complexity involved in the process. The CPCA attention mechanism is incorporated to zero in on key information and crucial regions, enhancing the algorithm?s detection precision. The BiFPN neck network structure is adopted to enhance cross-scale connections and information flow, improving the network's capability to identify objects of diverse sizes. Additionally, its efficient path and weight simplification mechanisms further reduce the overall computational burden of the network.

Herein lie the key advancements and contributions of this study:

(1) The introduction of PConv and EMA attention mechanisms within the feature extraction backbone aims to address the existing bottleneck, giving rise to the innovative C2F\_Faster\_EMA. By employing pointwise convolution along with effective feature aggregation strategies, this approach lowers the model?s parameter count without compromising detection accuracy, even in challenging environments.

(2) The SPPF\_CPCA multi-scale structure is designed by introducing the CPCA attention mechanism. Attention weights are dynamically allocated across channel and spatial dimensions. Multi-scale depth separable convolution modules are integrated to enhance the process. As a result, the detection accuracy in complex environments and target occlusion conditions is effectively improved.

(3) The incorporation of the BiFPN structure within the feature fusion segment bolsters cross-scale connections and information flow, thereby improving the network's capacity to identify objects of differing dimensions.

\section{Related Work}\label{sec2}

The YOLO series (You Only Look Once) stands out among various object detection algorithms due to its high speed, accuracy, and wide applicability. YOLOv8\citep{bib21}, built on YOLOv5, integrates improved features and refinements aimed at boosting both performance and versatility\citep{bib22}. The algorithm framework primarily consists of three parts: the backbone feature extraction network (Backbone), the neck feature fusion network (Neck), and the feature detection output layer (Head). The structure of YOLOv8 is illustrated in Fig.1.
\begin{figure}[h]
	\centering
	\includegraphics[width=0.9\linewidth]{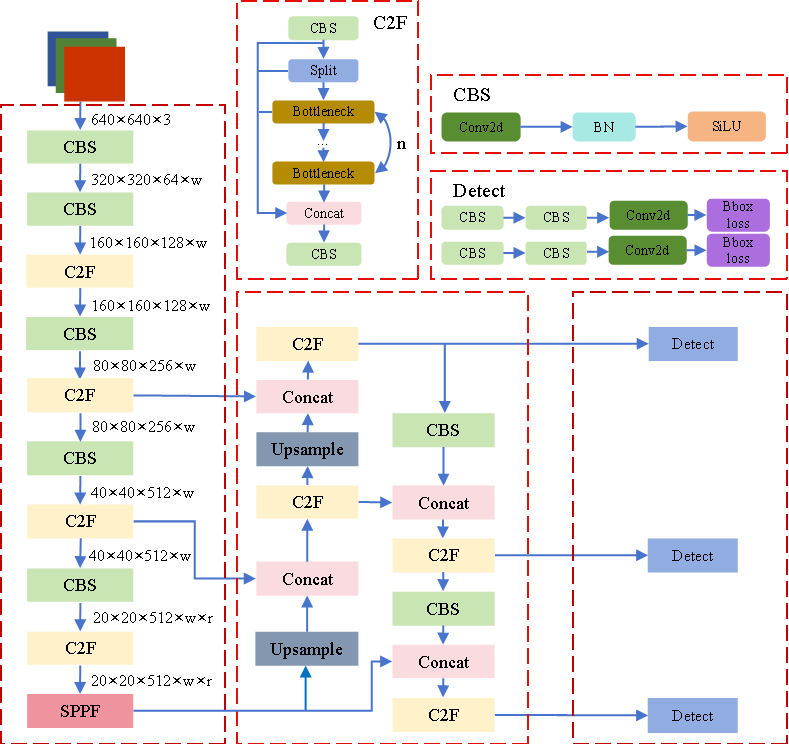}
	\caption{YOLOv8 network structure}
	\label{fig:v81}
\end{figure}

During detection tasks, the Backbone first performs feature extraction on the input image, using CSP (Cross Stage Partial) connections to enhance feature transmission. The backbone network consists of multiple CBS and C2F modules, along with an SPPF at the end. The CBS module is a composite module composed of Conv2d (two-dimensional convolution), BN (batch normalization), and SiLU (Sigmoid-Linear Unit) components. The convolutional layer extracts feature information from the input data by applying a set of learnable filters (also known as convolutional kernels or convolution matrices) through convolution operations. These filters possess varying feature extraction capabilities, effectively capturing edges, shapes, and other characteristics of the input data. To further elevate the network's expressive potential, the output is engaged through the non-linear activation function SiLU.

YOLOv8 integrates the C3 module from YOLOv5 and the ELAN module concept from YOLOv7\citep{bib23} to design the C2F module, replacing the C3 module. The C2F module employs gradient flow connections, enriching the feature extraction network's information flow while maintaining a lightweight structure. It consists of components such as CBS, Split, and Bottleneck. SPPF is a spatial pyramid pooling method based on SPP. It is implemented by serially applying multiple 5x5 max pooling layers. This method streamlines the number of parameters, leading to a reduction in computational demands. The SPPF module accepts feature maps as input, processes them through the CBS module, and performs max pooling downsampling operations. By using SPPF, the receptive field can be effectively expanded, allowing for the extraction of global contextual information and the fusion of features at different scales. Additionally, it sustains a minimal parameter count while keeping the computational burden light.

In the Neck network, the traditional FPN+PAN\citep{bib24,bib25} structure is still used. Compared to YOLOv7, YOLOv8's neck network removes convolution operations after upsampling, maintaining a lightweight structure.

The Head section presents a decoupled head architecture that employs distinct branches for classifying objects and refining bounding box locations. The bounding box regression employs DFL and CIoU\citep{bib26} structures, enhancing detection accuracy while accelerating model convergence. An anchor-free approach is also adopted, dynamically assigning positive and negative samples. BCE, Distribution Focal Loss, and CIOU loss functions are used in the loss computation.

\section{Methods}\label{sec3}

\subsection{Algorithm Structure}


The structure of the improved PEC-YOLO algorithm is shown in Fig.2. 
\begin{figure}[h]
	\centering
	\includegraphics[width=1\linewidth,height=10cm]{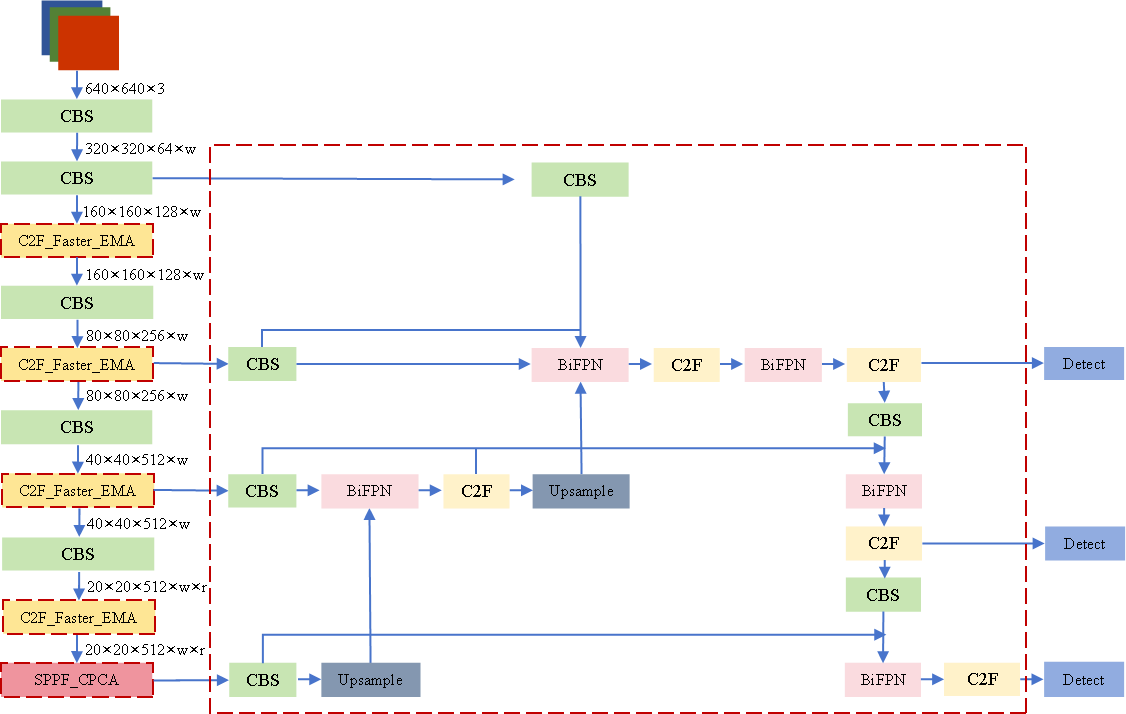}
	\caption{PEC-YOLO Network Structure}
	\label{fig:2}
\end{figure}

\subsection{PConv}\label{subsec2}

In the pursuit of creating swift and efficient neural networks, considerable attention has been devoted to minimizing the number of floating-point operations. However, reducing floating-point operations does not necessarily lead to a corresponding increase in computational speed. Frequent memory access during standard convolution operations results in lower floating-point operation efficiency. PConv (Partial Convolution) reduces computational redundancy and memory access, enabling more efficient extraction of spatial features\citep{bib27}. The working principle of PConv is illustrated in Fig.3(b). It leverages conventional convolution on only a subset of input channels for extracting spatial features, leaving the rest of the channels untouched..

\begin{figure}[H]
	\centering  
	\subfigbottomskip=2pt 
	\subfigcapskip=-5pt 
	\subfigure[Convolution]{
		\includegraphics[width=0.5\linewidth]{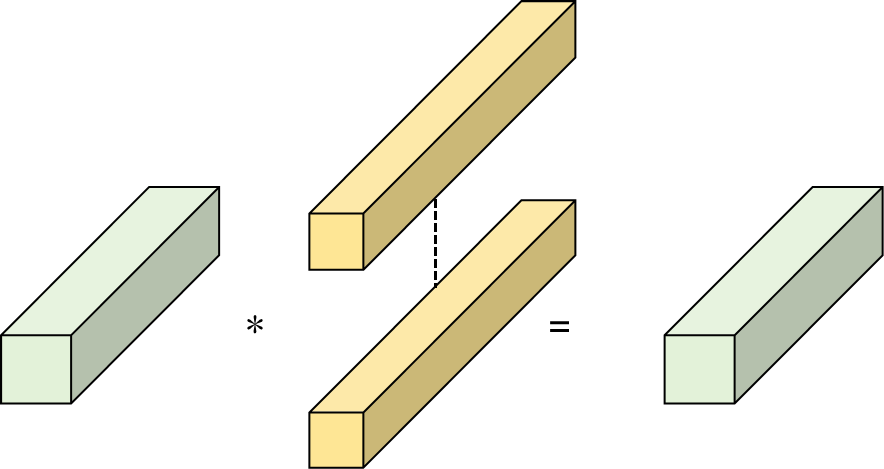}}
		\\
	\subfigure[Partial Convolution(PConv)]{
		\includegraphics[width=0.7\linewidth]{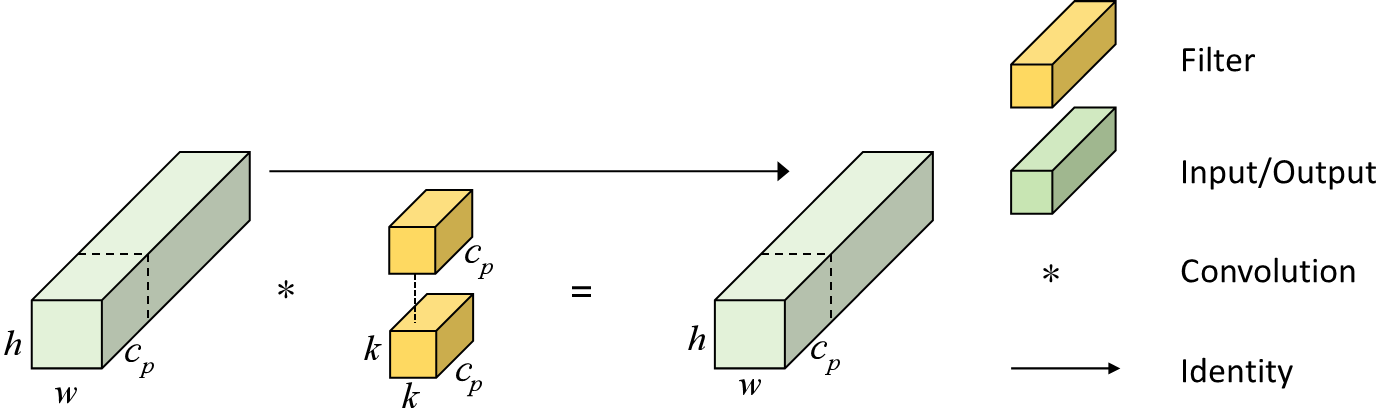}}
	\caption{Working Principle of PConv}
\end{figure}

For continuous or regular memory access, the first or last contiguous  channel is considered representative of the entire feature map for computation. It can be presumed that both the input and output feature maps have an equal number of channels. Therefore, the FLOPs of PConv are only $h\times w\times c_{p}^{2}$, for a typical ratio of $r=$1/4, the FLOPs of PConv are only 1/16 of those for regular Conv. Additionally, PConv has lower memory access requirements, that is $h\times w\times 2c_{p}+ k^{2}\times c_{p}^{2} \approx h\times w\times 2c_{p} $, for $r=$1/4, it requires only 1/4 of the memory access of regular Conv.

\subsection{EMA Attention Mechanism}\label{subsec2}

The purpose of the EMA(Efficient Multi-Scale Attention Module) module is to maintain the integrity of information across each channel, all while minimizing computational demands. It reformulates some of the channels into the batch dimension.S ubsequently, the channel dimension is grouped into several sub-features, which guarantees a balanced distribution of spatial semantic characteristics within each grouped feature\citep{bib28}. Alongside the encoding of global data to readjust the channel weights in each parallel pathway, the output features from both branches are subsequently consolidated. This aggregation is achieved through cross-dimensional interaction. This enhances the capability for multi-dimensional perception and multi-scale feature extraction. The configuration of the EMA attention module is illustrated in Fig. 4.
\begin{figure}[h]
	\centering
	\includegraphics[width=0.7\linewidth]{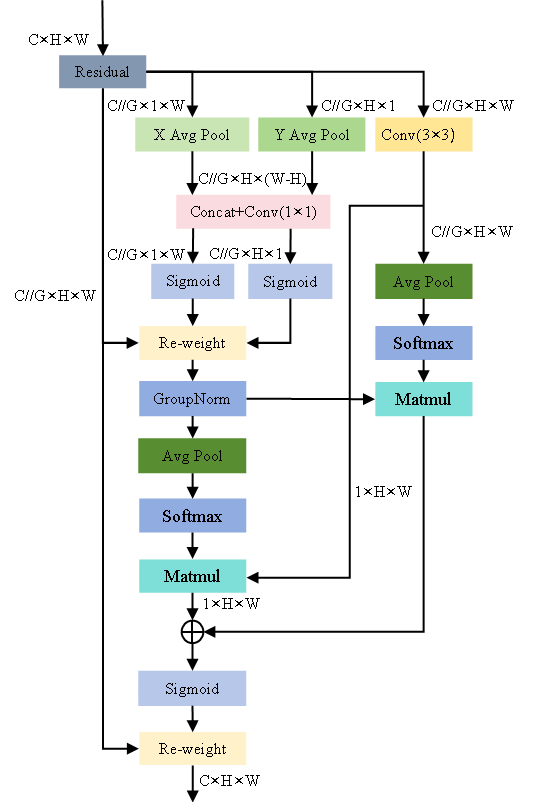}
	\caption{Structure of the EMA Attention Module}
	\label{fig:4}
\end{figure}

For any given input feature map $X \in      \mathbb{R}$, EMA divides the input along the channel dimension into G sub-features, that is $X = \left [ X_{0}, X_{i},\cdots ,X_{G- 1}    \right ] ,X\in \mathbb{R}^{C//G\times H\times W}$ which are used to learn different feature information. To capture the dependencies between all channels and reduce computational load, EMA models cross-channel feature interactions along the channel dimension. A different spatial dimension cross-space information aggregation method is also applied to achieve richer feature aggregation. Branch 1 reduces the channel count of the initial feature map and merges this information with that from the other branches. Branch 2, inspired by the CA attention mechanism, performs global average pooling on the feature map along both the height and width dimensions. To improve computational efficiency, the EMA attention mechanism employs two-dimensional average pooling, as shown in Equation (1). Branch 3 uses a $3\times 3$ convolution operation to process the feature map, effectively capturing cross-dimensional interactions and establishing connections between different dimensions with other branches. The three branches of the EMA attention mechanism integrate the advantages of channel attention and spatial attention. This approach captures global channel dependencies and local spatial features, thereby obtaining more comprehensive feature information across both channel and spatial dimensions.
\begin{align}Z_{c} & = \frac{1}{H\times W}\sum_{j}^{H}\sum_{i}^{W}x_{c}\left ( i,j \right )      \end{align}
Here, H and W represent the height and width of the feature map; $x_{c}$ denote the feature tensors of different channels.

\subsection{C2F\_Faster\_EMA Module}\label{subsec2}

This paper incorporates PConv along with the EMA attention mechanism to enhance the Bottleneck, effectively substituting the existing Bottleneck framework in C2F. A new C2F\_Faster\_EMA module has been constructed, which reduces the model's parameters and computational complexity. At the same time, it maintains the accuracy of target detection in complex environments. The specific improved Bottleneck structure is shown in Fig.5.
\begin{figure}[!ht]
	\centering
	\begin{minipage}[b]{0.45\textwidth}
		\centering
		\subfigure[Original Bottleneck]{\includegraphics[width=2.5cm]{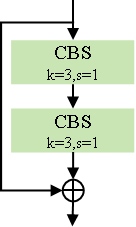}}
	\end{minipage}
	\begin{minipage}[b]{0.45\textwidth}
		\centering
		\subfigure[Improved Bottleneck]{\includegraphics[width=0.45\linewidth]{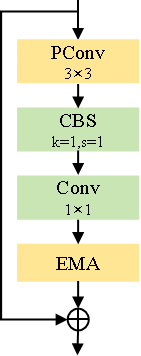}}
	\end{minipage}
	\caption{Bottleneck Structure Diagram}
\end{figure}

At the input end, PConv applies regular Conv to a portion of the input channels for spatial feature extraction, while keeping the remaining channels unchanged. The convolution results are then concatenated with the aforementioned non-convoluted channels, reducing redundant information. This is followed by a CBS module, which doubles the number of channels in the output feature map, maintaining feature diversity and achieving lower latency. Then, a $1\times 1$ convolution is applied to reduce the dimensionality, ensuring the number of channels matches the input. The resultant output is subsequently channeled into the EMA attention module, which captures global information to adjust the channel weights across each parallel branch. This achieves cross-dimensional aggregation of output features, enhancing multi-dimensional perception and multi-scale feature extraction capabilities.
\subsection{CPCA}\label{subsec2}
The CPCA (Channel Prior Convolution Attention) method dynamically allocates attention weights across channel and spatial dimensions. Utilizing a multi-scale depthwise separable convolution module to establish spatial attention allows for the effective extraction of spatial relationships, all while preserving existing channels. This method focuses on key information channels and regions\citep{bib29}.

Unlike CBAM(Convolutional Block Attention Module), CPCA employs a depthwise convolution module to establish the spatial attention element. The depthwise convolution module employs elongated convolution kernels of various sizes to capture the spatial relationships between pixels. By utilizing these multi-scale, bar-shaped convolution kernels, it manages to extract crucial information while simultaneously minimizing computational demands.  Initially, the channel attention module is implemented to generate the channel attention map. Following this, the depthwise convolution module progressively identifies the crucial spatial areas within each channel, generating spatial attention maps that are distributed dynamically for each channel. These attention maps closely mirror the genuine feature distribution, improving the segmentation performance of the network.

The CPCA employs a cohesive framework that systematically arranges CA(channel attention) followed by SA(spatial attention), as illustrated in Fig. 6. It utilizes techniques like average and max pooling to consolidate the spatial data from the feature map. The spatial data is subsequently channeled through a shared MLP(multi-layer perceptron) and integrated into the channel attention map. Channel priors are derived by multiplying the input features element-wise with the channel attention map. These prior channels are then fed into the depthwise convolution module, which produces the spatial attention map. The convolution module takes in the spatial attention map to facilitate channel mixing. Ultimately, the refined features are generated by carrying out element-wise multiplication between the channel-mixed results and the channel priors. This mixing process plays a crucial role in improving the feature representation.
\begin{figure}
	\centering
	\includegraphics[width=1\linewidth]{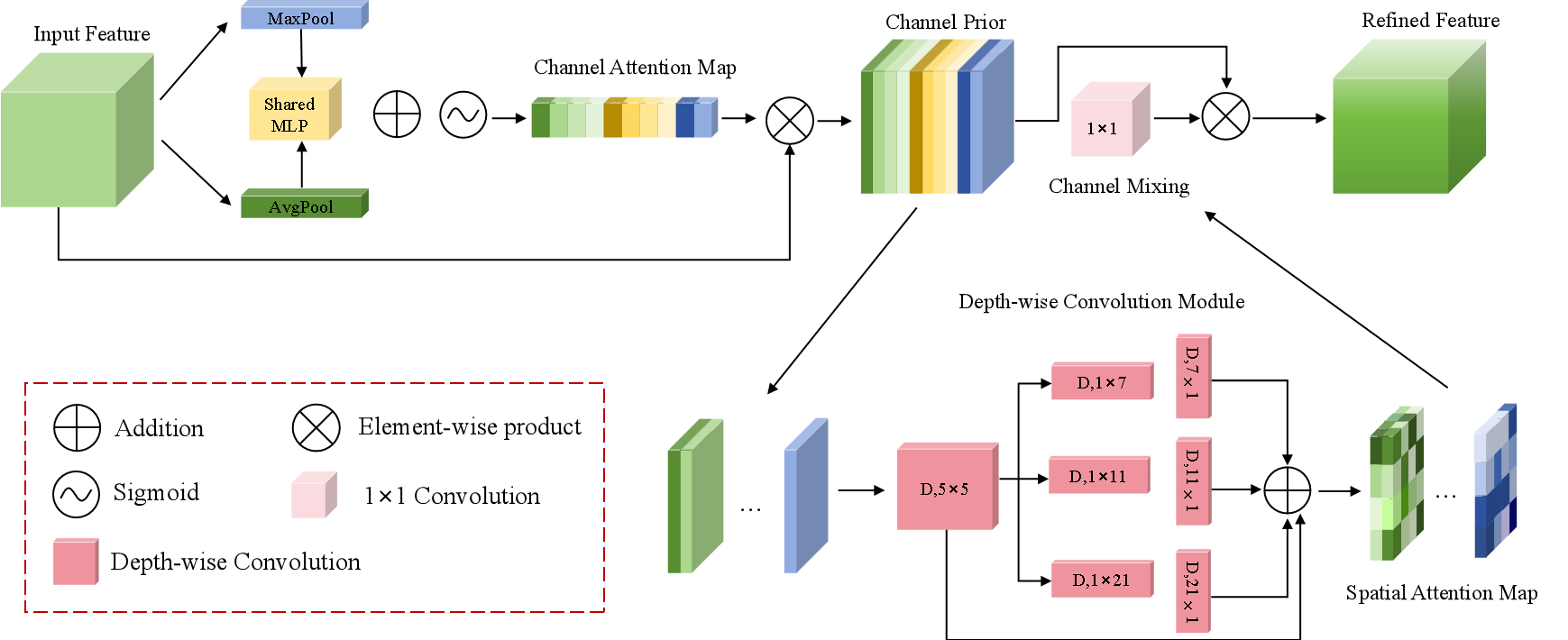}
	\caption{Overall Structure of Channel Prior Convolution Attention}
	\label{fig:6}
\end{figure}
Starting with an intermediate feature map $F\in \mathbb{R}^{C\times H\times W}$, CA initially generates a 1D channel attention map $M_{c}\in \mathbb{R}^{C\times 1\times 1}$. The attention map is subsequently element-wise multiplied with the input feature $F$, effectively extending the channel attention values across the spatial dimension to produce enhanced features that incorporate channel attention $F_{c} \in \mathbb{R}^{C\times H\times W}$. The 3D spatial attention map $M_{s} \in \mathbb{R}^{C\times H\times W}$ is generated through the SA process $F_{c}$ . The final output features $\hat{F} \in \mathbb{R}^{C\times H\times W}$ are derived by performing element-wise multiplication of the previous results. The overall attention mechanism can be summarized as follows:\newpage\begin{align}F_{c} & = CA\left ( F \right )  \otimes F\end{align}\begin{align}\hat{F}  & = SA\left ( F_{c}  \right )  \otimes F_{c} \end{align}
Here, $\otimes$ denotes element-wise multiplication.

Channel Attention: Following the CBAM approach, average pooling and max pooling operations are used to aggregate spatial information from the feature map, which is then input into a shared MLP. The results from the shared MLP are then aggregated using element-wise summation to produce the final channel attention map. To minimize parameter overhead, the shared MLP is designed with a single hidden layer, where the size of the hidden activation is configured to $\mathbb{R}^{\frac{C}{r}\times 1\times 1 }$, with a reduction ratio denoted as $r$. The calculation method for channel attention is as follows:
\begin{align}CA\left ( F \right ) & = \sigma \left ( MLP\left ( AvgPool\left ( F \right )  \right ) +  MLP\left ( MaxPool\left ( F \right )  \right )  \right ) \end{align}
Here, $\sigma$ denotes the sigmoid function.

Spatial Attention: Depthwise separable convolution is utilized to capture the spatial relationships between features. This configuration effectively preserves inter-channel relationships while streamlining computational demands. Additionally, a multi-scale framework is utilized to improve the convolution operations' capacity to grasp spatial relationships. Finally, a $1\times 1$ convolution is used for channel mixing. The calculation method for spatial attention is as follows:
\begin{equation}
	SA\left ( F \right ) = Conv_{1\times 1}\left (  {\textstyle \sum_{i = 0}^{3}} Branch_{i}\left ( DwConv\left ( F \right )  \right )   \right )
\end{equation}
Here, $DwConv$ denotes depthwise convolution; $Branch_{i},i\in \left \{ 0,1,2,3 \right \}$ denotes the i-th branch; $Branch_{0}$ denotes the identity connection.
\subsection{SPPF\_CPCA Module}\label{subsec2}

To enhance the focus on target regions and address detection issues in complex environments and under severe occlusion, the CPCA attention mechanism is used to improve the SPPF module. This results in the design of the SPPF\_CPCA structure, as shown in Fig.7. First, the input feature map is processed by the CBS module and then sequentially passed through three $5\times 5$ max pooling layers for downsampling. This further captures target feature information and achieves multi-scale perception. Then, the different downsampling results are concatenated and passed through the CPCA attention mechanism. This effectively extracts spatial relationships and retains prior channel information, thereby focusing on key information and important regions. As a result, it becomes easier to locate fine-grained features, significantly enhancing the overall performance of the model.
\begin{figure}[h]
	\centering
	\includegraphics[width=1\linewidth]{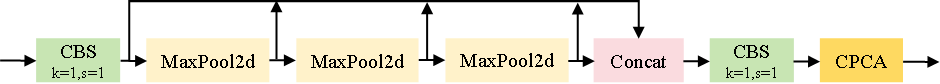}
	\caption{Structure of the SPPF\_CPCA Module}
	\label{fig:7}
\end{figure}
\subsection{Improved Neck Network Structure}\label{subsec2}
The FPN (Feature Pyramid Network) structure is shown in Fig.8(a). FPN combines both high-level and low-level semantic features through a top-down approach, effectively enhancing feature localization. However, in deep neural networks, the path for transferring shallow features to deep features is generally long. During downsampling, most of the target information may have already been lost. This method does not effectively achieve feature fusion. The PANet (Path Aggregation Network) structure is shown in Fig.8(b). It adds an additional bottom-up pathway to the FPN, directly fusing the lower-level spatial features upwards, thereby retaining more shallow features. PANet optimizes the feature fusion method of the FPN network to a certain extent, enhancing the object detection performance. However, it also increases the network's parameter count and computational load. To address the issues present in PANet, BiFPN (Bidirectional Feature Pyramid Network) is introduced, as shown in Fig.8(c)\citep{bib30}.
	\begin{figure}[ht]
	\centering
	\begin{minipage}[b]{0.32\textwidth}
		\centering
		\subfigure[FPN Structure]{\includegraphics[width=3cm, height=5.4cm]{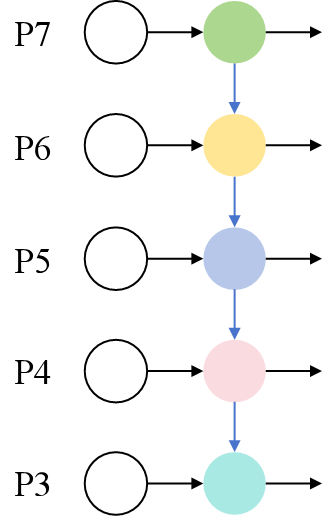}}
	\end{minipage}
	\begin{minipage}[b]{0.32\textwidth}
		\centering
		\subfigure[PANet Structure]{\includegraphics[width=4cm, height=5.4cm]{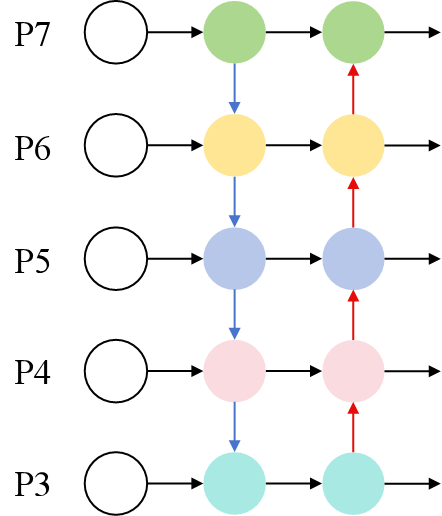}}
	\end{minipage}
	\begin{minipage}[b]{0.32\textwidth}
		\centering
		\subfigure[BiFPN Structure]{\includegraphics[width=4cm, height=6.2cm]{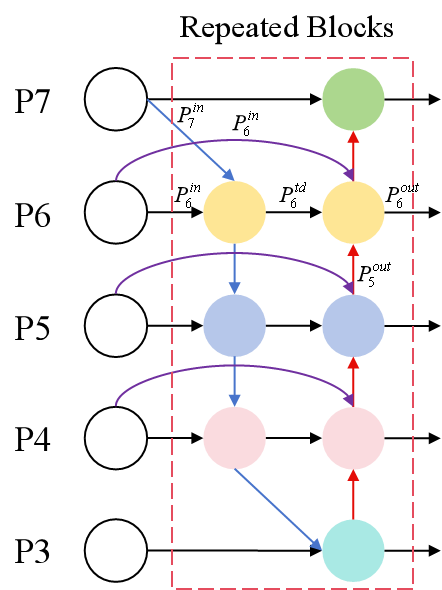}}
	\end{minipage}
	\caption{Diagrams of Three Feature Pyramid Structures}
\end{figure}
The key improvements include: (1) Removing nodes with a single input edge to enhance contribution; (2) Introducing additional connections when the input and output nodes are aligned at the same hierarchy, to enable a richer feature fusion without escalating the computational expense; (3) Treating bidirectional paths as a single unit, where a basic network layer is repeatedly stacked multiple times to fuse more advanced features; (4) In weighted feature map fusion, BiFPN fine-tunes the impact of each scale feature by assigning specific weights to them.

Taking the characteristics of the intermediate layer for instance, the two fused features of BiFPN at level 6 are described. Here, the intermediate feature from level 2 in the top-down pathway is denoted as ${P}_{6}^{td}$, while the output feature from level 6 in the bottom-up pathway is represented as ${P}_{6}^{out}$, as seen in Equations (6) and (7). The construction methods for other features are similar. It is important to highlight that BiFPN employs depthwise separable convolutions for combining features, incorporating batch normalization and activation following each convolutional layer.
\begin{equation}
{P}_{6}^{td}=Conv \left ( \frac{{\omega }_{1}{P}_{6}^{in}+{\omega }_{2}Resize\left ({P}_{7}^{in} \right )}{{\omega }_{1}+{\omega }_{2}+\varepsilon }\right )
\end{equation}
\begin{equation}
	{P}_{6}^{out}=Conv \left ( \frac{{{{\omega }_{1}}^{\prime}}{P}_{6}^{in}+{{\omega }_{2}}^{\prime}{P}_{6}^{td}+{{\omega }_{3}}^{\prime}Resize\left ({P}_{5}^{in} \right )}{{{\omega }_{1}}^{\prime}+{{\omega }_{2}}^{\prime}+{{\omega }_{3}}^{\prime}+\varepsilon }\right )
\end{equation}
Here, $Resize$ denotes the upsampling or downsampling operations; $\omega$ represents the learned parameters used to distinguish the importance of different features during the feature fusion process.
\section{Experiment and Analysis}\label{sec4}
\subsection{ Dataset}\label{subsec2}
The dataset used in this paper comes from the Alibaba Tianchi Competition: Guangdong Power Grid Intelligent On-site Operation Challenge. There are four types of detection objects: supervisor armbands, people in off-ground status, people in on-ground status, and people wearing safety belts. The dataset contains a total of 2546 images, with the distribution of label numbers shown in Table 1. The data annotation format is converted to txt format, and the dataset is randomly divided into training, validation, and test sets in a ratio of 7:1:2. The training set is used to train the parameters of the object detection algorithm to obtain the training weights for this dataset. The training dataset is utilized to tune the parameters of the object detection algorithm, which helps in acquiring the relevant weights for this specific dataset. Meanwhile, the validation dataset plays a crucial role in overseeing the training process, helping to mitigate the risk of overfitting. The test dataset serves to assess both the efficacy of the training process and the algorithm's performance. In this experiment, the training set contains 1782 images, the validation set contains 255 images, and the test set contains 509 images.
\begin{table}[h]
	\caption{Number of Labels in the Dataset}\label{tab1}%
	\begin{tabular}{@{}llll@{}}
		\toprule
		Label Name & Meaning  & Count\\
		\midrule
		Badge    & supervisor armbands  & 673   \\
		Offground   & people in off-ground status   & 2477   \\
		Ground  & people in on-ground status   & 2257   \\
		Safebelt  & people wearing safety belts   & 1747   \\
		\botrule
	\end{tabular}
\end{table}

\subsection{ Experimental Setting}\label{subsec2}
Throughout the experiments, model construction, training, and testing are conducted using Pytorch version 2.1.1. The Python version used is 3.8.0, and the CUDA version is 12.1. The GPU utilized is an NVIDIA GeForce RTX 3090. The dimensions of the input images for the model are configured to be $640\times640\times3$. The thresholds for non-maximum suppression and IoU (Intersection over Union) are set to 0.3 and 0.5, respectively. The optimization strategy employs the SGD optimizer for gradient descent. The batch size is set to 32, the number of epochs to 400, and the initial learning rate to 0.001.
\subsection{ Evaluation Index}\label{subsec2}
In this experiment, the precision, recall, mean average precision (mAP), detection speed (FPS), and parameter quantity are employed to assess the overall recognition performance of the PEC-YOLO model across various categories within the power sector. P (Precision) and R (Recall) can be expressed as:
\begin{equation}
	P=\frac{{T}_{P}}{{T}_{P}+{F}_{P}}
\end{equation}
\begin{equation}
	R=\frac{{T}_{P}}{{T}_{P}+{F}_{N}}
\end{equation}

Here, ${T}_{P}$ refers to the count of true positive samples that are predicted as positive  and are indeed accurate. ${F}_{P}$ denotes the count of false positives samples, which are incorrectly classified as positive when they are actually negative. Conversely, ${F}_{N}$ indicates the number of false negatives, which refers to samples that are mistakenly labeled as negative yet are true positives. Precision (P) and Recall (R) can sometimes be at odds with each other. As a result, the area beneath the P-R curve is typically calculated as the AP (Average Precision) metric to measure the performance of object detection. It can be expressed as:
\begin{equation}
	A_{P}  =  \int_{0}^{1}P\left ( R \right )dR
\end{equation}
	
The average of the AP values across various categories ${M}_{AP}$ can be represented as:
\begin{equation}
	M_{AP}=\frac{\sum\limits_{i=1}^{N}A_{Pi}}{N} 
\end{equation}
In the formula, the subscript $i$ denotes the category index; $N$ represents the total number of categories in the dataset. The evaluation metrics used in this experiment are $M_{AP}\left ( 0.5 \right )$ and $M_{AP}\left ( 0.5:0.95\right )$. Here, $M_{AP}\left ( 0.5 \right )$ indicates that a detection is considered successful if the IoU threshold between the predicted result and the actual target is greater than or equal to 0.5; $M_{AP}\left ( 0.5:0.95\right )$ represents the average value of mAP across different IoU thresholds, ranging from 0.5 to 0.95, with a step size of 0.05.

FPS represents the number of images detected per second. It reflects the operational speed of the detection model. The calculation formula is as follows:
\begin{equation}
	FPS=\frac{f_{n} }{T} 
\end{equation}
Here, $f_{n}$ represents the total number of images, and $T$ represents the total time taken.

\subsection{ Training Experiment Analysis}\label{subsec2}
Through validation on the validation set, the evaluation metrics and loss value curves for the original YOLOv8 algorithm and the PEC-YOLO algorithm are obtained. These are shown in Fig.9 and Fig.10, respectively. After 400 epochs of training, both the improved algorithm and the original algorithm gradually stabilize, achieving optimal detection accuracy and minimal loss. Fig.9 indicates that, in contrast to the original algorithm, the enhanced algorithm exhibits a minor decline in Precision, while it shows gains in Recall, mAP@0.5, and mAP@0.5:0.95. It is important to highlight that, unlike Precision, mAP offers a more holistic assessment metric that effectively balances the interplay between Precision and Recall. When evaluating the performance of an algorithm model, placing greater emphasis on the mAP value is a more reasonable choice. It is shown in Fig.10 that the improved algorithm has lower loss compared to the original algorithm, confirming the feasibility of the improvement strategy.
\begin{figure}[H]
	\centering
	
	\begin{minipage}[b]{0.45\textwidth}
		\centering
		\subfigure[Precision ]{
			\includegraphics[width=5cm, height=5cm]{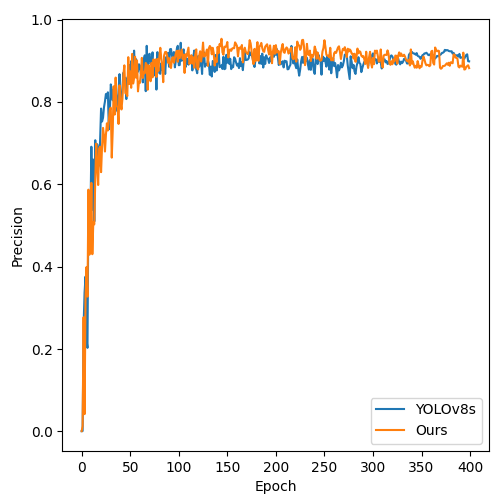}
		}
	\end{minipage}
	\begin{minipage}[b]{0.45\textwidth}
		\centering
		\subfigure[Recall]{
			\includegraphics[width=5cm, height=5.01cm]{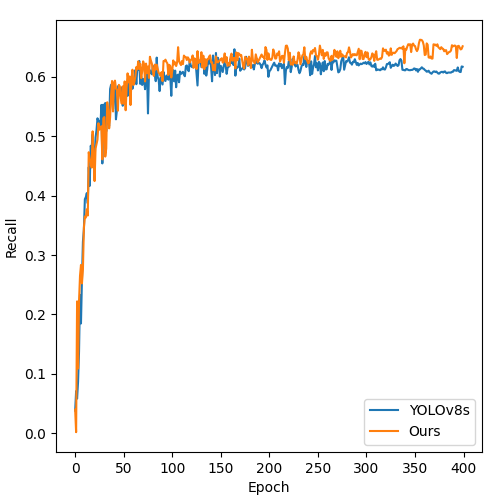}
		}
	\end{minipage}
	
	\vspace{0.1cm} 
	
	\begin{minipage}[b]{0.45\textwidth}
		\centering
		\subfigure[mAP@.5]{
			\includegraphics[width=5cm, height=5cm]{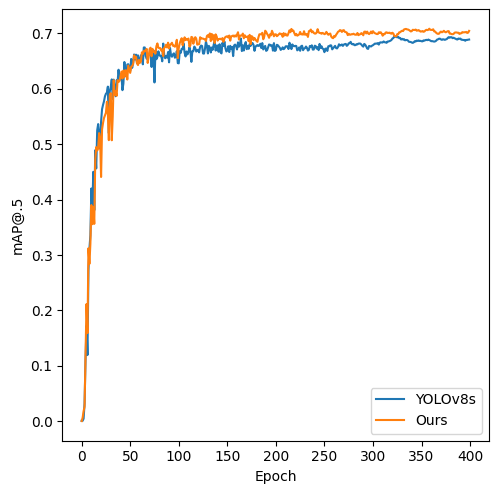}
		}
	\end{minipage}
	\begin{minipage}[b]{0.45\textwidth}
		\centering
		\subfigure[mAP@.5:.95]{
			\includegraphics[width=5cm, height=5.01cm]{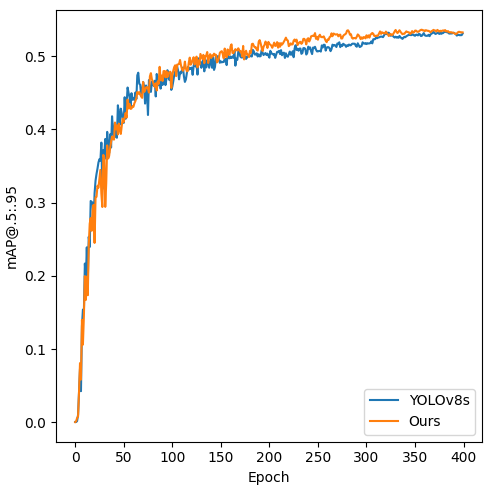}
		}
	\end{minipage}
	
	\caption{Comparison of Evaluation Metric Curves}
\end{figure}

\begin{figure}
\centering
\begin{minipage}[b]{0.45\textwidth}
			\centering
			\subfigure[Box\_loss]{
		\includegraphics[width=5cm, height=5cm]{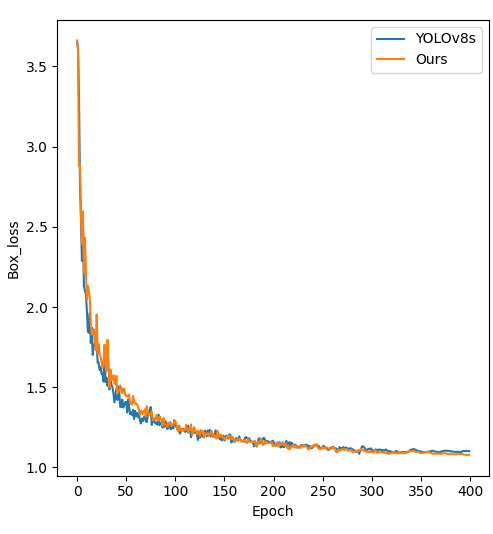} }
\end{minipage}
\begin{minipage}[b]{0.45\textwidth}
		\centering
		\subfigure[Dfl\_loss]{
		\includegraphics[width=5cm, height=5.07cm]{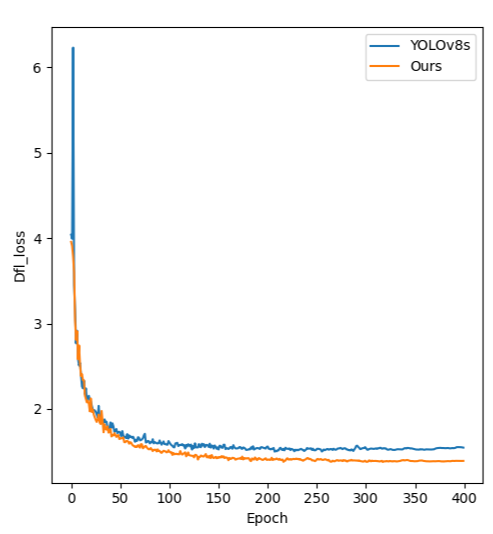}	}
\end{minipage}
\vspace{0.1cm}
\begin{minipage}[b]{0.45\textwidth}
		\centering
		\subfigure[Cls\_loss]{
		\includegraphics[width=5cm, height=5cm]{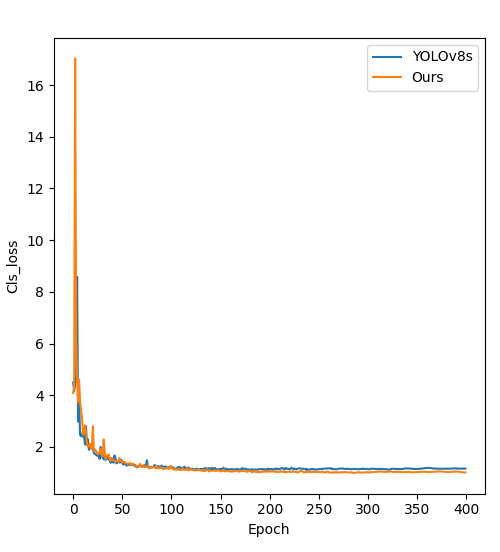} 
	}
\end{minipage}
	\caption{ Comparison of Loss Value Curves} 
\end{figure}

\subsection{Ablation Experiment Analysis}\label{subsec2}
To better evaluate the effectiveness of the improved algorithm, ablation experiments were conducted under the same conditions. The details are shown in Table 2. In Table 2, the first column represents the experiment number, the second column compares the ablation experiment with the original experiment algorithm, and the third to eighth columns present the performance metrics for evaluating the algorithm. After replacing the C2F module in the backbone network, the model's detection speed increases by nearly 500 frames, and the parameter quantity decreases by 12.85\%. Precision, Recall, and mAP@0.5 increase by 0.5\%, 1.6\%, and 0.2\%, respectively, indicating that the improved module is lightweight and achieves a certain improvement in accuracy. However, mAP@0.5:0.95 decreases by 0.8\%. It is important to note that compared to mAP@0.5:0.95, mAP@0.5 is a more critical metric for evaluating the detection accuracy of algorithm models. mAP@0.5 considers the model's performance at a more lenient IoU threshold, which is more practically significant for object detection tasks. Its improvement indicates that the algorithm can accurately detect as many targets as possible. Overall, the improvement of the C2F module ensures the model's lightweight nature while enhancing the average detection accuracy to a certain extent. Subsequently, the SPPF module is improved using CPCA, leading to a further enhancement in algorithm accuracy. Compared to Experiment 2, Experiment 3 shows a 2\% increase in Recall, and mAP@0.5 and mAP@0.5:0.95 increase by 1.6\% and 1.7\%, respectively. However, there is a certain degree of decline in detection speed, and the parameter quantity has increased to some extent. The attention mechanism, while increasing algorithm accuracy, also results in an increase in the number of parameters. This is a common drawback of attention mechanisms. After modifying the feature fusion part to a BiFPN structure, compared to Experiment 3, the parameter quantity decreases by 37.04\%, and detection speed increases by approximately 460 frames. Simultaneously, the detection accuracy further improves, with mAP@0.5 and mAP@0.5:0.95 increasing to 79.9\% and 63.4\%, respectively. This fully utilizes the redundant feature information.

Ultimately, the PEC-YOLO algorithm proposed in this paper shows a 4.9\% improvement in Recall compared to the original algorithm, reducing missed detections. The mAP@0.5 and mAP@0.5:0.95 increase by 2.7\% and 1.1\%, respectively, significantly enhancing detection performance. The overall detection speed improves by nearly 400 frames, and the number of parameters decreases by 42.59\%, making the model faster and more lightweight.
\begin{table*}[!h]
	\caption{Ablation Experiment Results}\label{tab2}
	\renewcommand{\arraystretch}{2.5}
	\resizebox{\textwidth}{!}{
	\begin{tabular}{@{}llllllll@{}}
		\toprule
		Serial Number & Model  & Precision
		(\%) & Recall
		(\%) & mAP@.5
		(\%) & mAP@.5:.95
		(\%)  & Detection Speed
		(FPS) & Parameter
		($10^{6}$)\\
		\midrule
		1    & YOLOv8s   & 92.4  & 60.7 & 77.2   & 62.3   & 125.0 & 11.13 \\
		2    & YOLOv8s+C2F\_Faster\_EMA   & 92.9 & 62.3 & 77.4    & 61.5   & 625.0  & 9.70 \\
		3    & \parbox{1.8cm}{YOLOv8s+C2F\_Faster\_EMA\\+SPPF\_CPCA}   & 91.4  & 64.3 & 79.0    & 63.2   & 63.7  & 10.15 \\
		4    & \parbox{1.8cm}{YOLOv8s+C2F\_Faster\_EMA\\+SPPF\_CPCA+BiFPN}   & 90.5  & \textbf{65.6} & \textbf{79.9}    & \textbf{63.4}  & \textbf{526.3}  & \textbf{6.39}\\
		\botrule
	\end{tabular}}
\end{table*}
\FloatBarrier 

\subsection{Comparative Experiment Analysis}\label{subsec2}
The capabilities of PEC-YOLO in detecting power fields are confirmed through a comparison with current object detection algorithms. The methods compared include one-stage methods (YOLOv5s, YOLOv7, YOLOv8s, SSD\citep{bib31}, CenterNet\citep{bib32}) and the two-stage method (Faster R-CNN\citep{bib33}). In this experiment, the training was carried out using the training set, adhering to consistent experimental conditions, and followed the suggested training parameters of the comparative techniques. The experimental results are shown in Table 3. Analyzing the data presented in the table reveals that PEC-YOLO has enhanced its performance over the original YOLOv8 network. The recall rate, mAP@0.5, and mAP@0.5:0.95 have increased by 4.9\%, 2.7\%, and 1.1\%, respectively, indicating a certain improvement in detection accuracy. Meanwhile, the improved PEC-YOLO algorithm model shows a nearly 400-frame increase in detection speed. Additionally, the network parameters are reduced by 42.59\%, enhancing its suitability for mobile device implementation. The above results demonstrate the superiority and effectiveness of the proposed YOLOv8-based improvement method in power field detection. Compared to other methods, PEC-YOLO performs better. This is primarily due to the robust design of the YOLOv8 model and the specific improvements proposed for the power field environment.
\begin{table*}[!h]
	\caption{Comparative Experiment Results}\label{tab2}
	\resizebox{\textwidth}{!}{
		\begin{tabular}{@{}lllllll@{}}
			\toprule
			Comparison Method & Precision
			(\%)  & Recall
			(\%) & mAP@.5
			(\%) & mAP@.5:.95
			(\%) & Detection Speed
			(FPS) & Parameter
			($10^{6}$)\\
			\midrule
			Faster R-CNN   & 60.9   & 56.2  & 53.3 & 25.7   & 43.3   & 136.75 \\
			SSD  & 89.8  & 58.4 & 67.5 & 39.4    & 71.3   & 24.15 \\
			Centernet  & \textbf{97.8} & 47.1  & 69.1 & 38.7    & 110.8  & 32.66 \\
			YOLOv5s  & 87.7  & \textbf{66.8}  & 71.2 & 55.5   & 312.5  & 7.02 \\
			YOLOv7  & 93.9  & 64.5  & 71.7 & 59.5   & 68.0  & 36.49 \\
			YOLOv8s  & 92.4 & 60.7  & 77.2 & 62.3   & 125.0  & 11.13 \\
			PEC-YOLO  & 90.5 & 65.6  & \textbf{79.9} & \textbf{63.4}   & \textbf{526.3}  & \textbf{6.38} \\
			\botrule
	\end{tabular}}
\end{table*}

\subsection{ Detection Results Analysis}\label{subsec2}

To more intuitively verify the practical detection effect of the improved algorithm, Table 4 provides examples of detection results on the test set before and after the improvement. In the first row, under conditions of a relatively simple background and partial occlusion of the target, the improved model can more accurately detect whether a safety belt is worn. This model addresses the missed detection issue of YOLOv8s, achieving an accuracy of 0.80. In the second row, under the condition of a more complex outdoor detection background, the improved model can effectively detect the off-ground status and safety belt wearing status of the worker above. In the third row, under the condition of a more complex outdoor detection background, the improved model can reduce the occurrence of false detections. In the fourth row, under the condition of a more complex outdoor detection background, the detection accuracy for the off-ground status increased from 0.69 to 0.81. In the fifth row, under a typical complex background, the improved model can reduce false detections in the lower-left corner. The detection accuracy for safety belt wearing increased from 0.49 to 0.87, and the detection accuracy for the off-ground status increased from 0.40 to 0.93. In the sixth row, under a typical complex background, the improved model can correctly detect the previously missed targets of safety belt wearing and on-ground status. This reduces the occurrence of missed detections. In the seventh row, under typical complex environments with high target variance, the improved model can effectively detect the previously missed target of the supervisor armband in the lower-left corner. Additionally, the detection accuracy for people in the on-ground status increased from 0.40 to 0.93, and for people in the off-ground status increased from 0.69 to 0.91.

Finally, the attention regions of the images are visualized using HiResCAM\citep{bib34}  heatmaps for comparison, as shown in Table 5. The heatmap visualization reveals that the areas of focus correspond to the centers of the detection boxes. Red indicates the central, most focused areas, with decreasing attention as it spreads outward. This shows that the model is focusing on the correct detection areas. Additionally, red circles highlight the comparison between the detections before and after the improvements.
\begin{table}[!h]
	\caption{Detection Results of Test Set Images}\label{tab4}%
	\begin{tabular}{@{}ll@{}}
			\toprule
			YOLOv8s & PEC-YOLO\\
			\midrule
			\includegraphics[width=3.5cm]{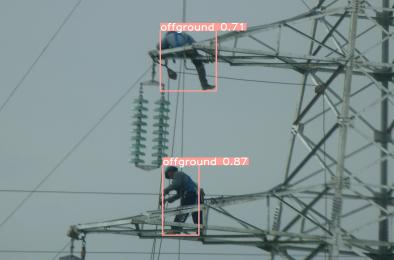}  & \includegraphics[width=3.5cm]{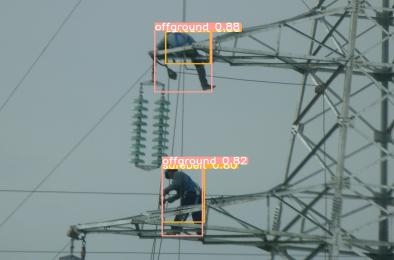}  \\
			\includegraphics[width=3.5cm]{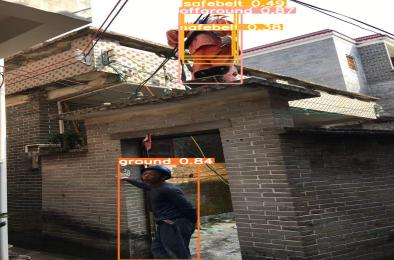}  & \includegraphics[width=3.5cm]{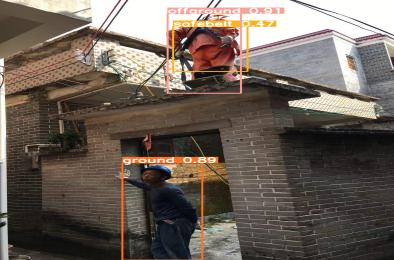}  \\
			\includegraphics[width=3.5cm]{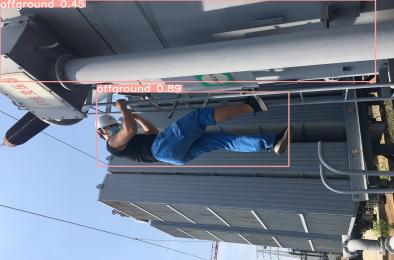}  & \includegraphics[width=3.5cm]{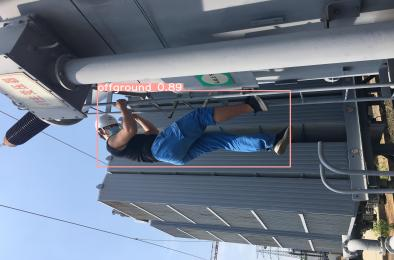}    \\
			\includegraphics[width=3.5cm]{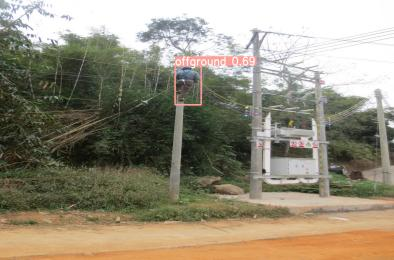}  & \includegraphics[width=3.5cm]{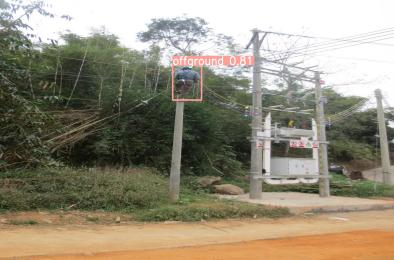}  \\
			\includegraphics[width=3.5cm]{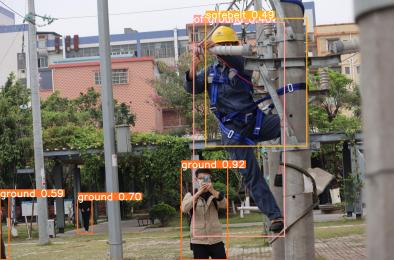}  & \includegraphics[width=3.5cm]{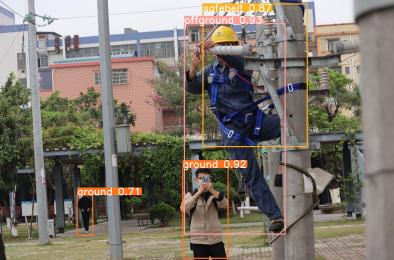}  \\
			\includegraphics[width=3.5cm]{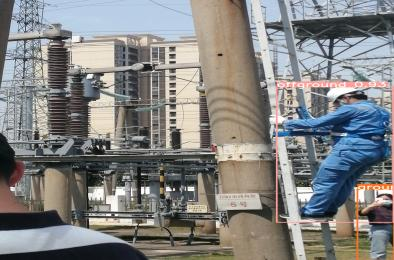}  & \includegraphics[width=3.5cm]{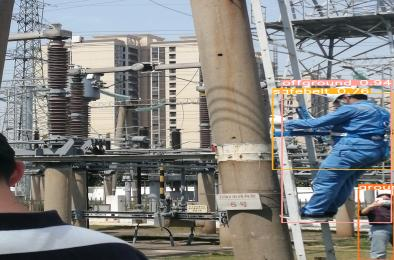}  \\
			\includegraphics[width=3.5cm]{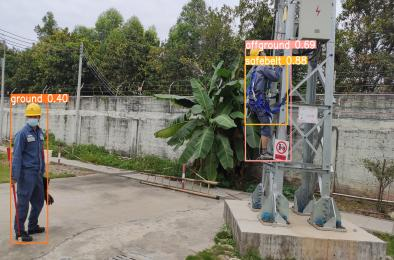}  & \includegraphics[width=3.5cm]{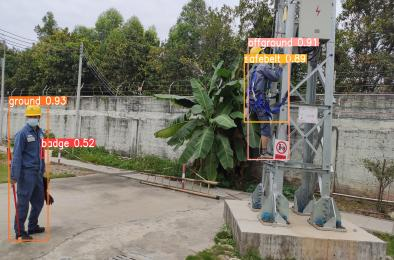}  \\
			\botrule
	\end{tabular}
\end{table}

In the first and second rows of images in Table 5, the improved model more effectively detects the wearing status of the supervisory armband. Additionally, it provides more accurate detection of both off-ground and on-ground states. In the third row of images, the improved model effectively detects off-ground targets that are missed by the original model. In the fourth row of images, the improved model provides more accurate detection areas for off-ground targets and safety belts. In the fifth row of images, the improved model effectively detects off-ground targets while eliminating false detections made by the original model. In summary, the improved PEC-YOLO model can more accurately detect the status of electrical site workers in complex environments, with partial target occlusion, and high target variance.
\begin{table}[!h]
	\caption{The heatmap visualization results based on HiResCAM}\label{tab4}%
	\begin{tabular}{@{}ll@{}}
		\toprule
		YOLOv8s & PEC-YOLO\\
		\midrule
		\includegraphics[width=3.5cm]{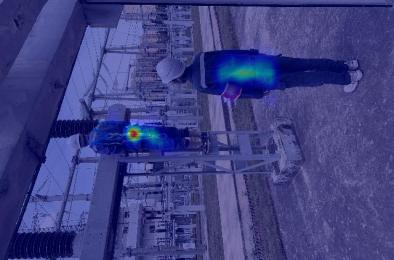}  & \includegraphics[width=3.5cm]{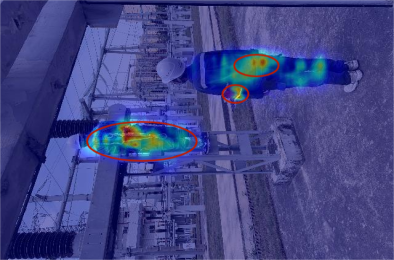}  \\
		\includegraphics[width=3.5cm]{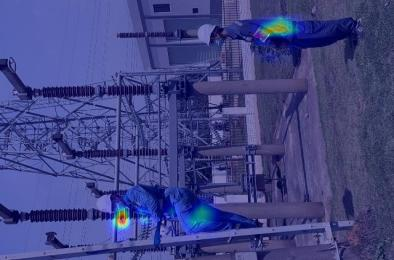}  & \includegraphics[width=3.5cm]{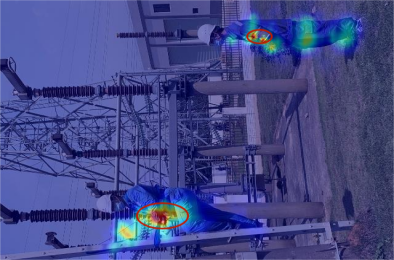}  \\
		\includegraphics[width=3.5cm]{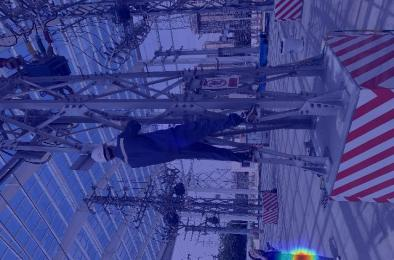}  & \includegraphics[width=3.5cm]{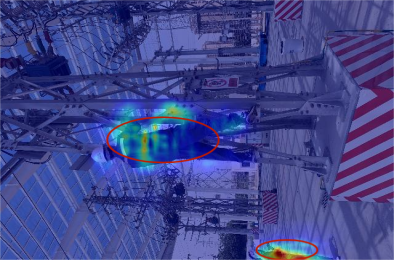}    \\
		\includegraphics[width=3.5cm]{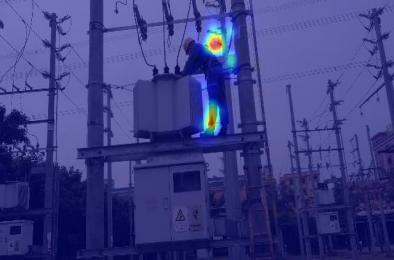}  & \includegraphics[width=3.5cm]{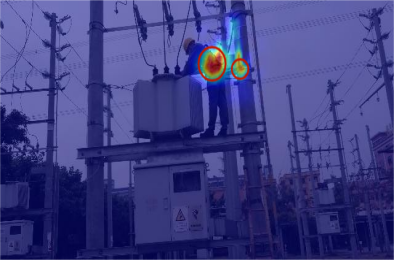}  \\
		\includegraphics[width=3.5cm]{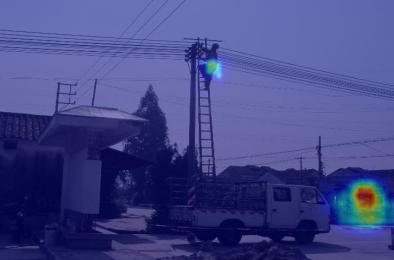}  & \includegraphics[width=3.5cm]{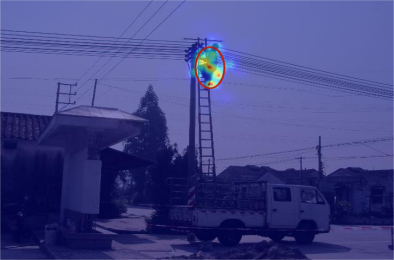}  \\
		\botrule
	\end{tabular}
\end{table}
\FloatBarrier

\section{Conclusion}\label{sec13}
This paper addresses the issues of complex backgrounds, target occlusion, and large target variance in power site operations, while also considering edge device deployment. An improved PEC-YOLO algorithm is proposed. 

The key improvement lies in the formation of a new C2F\_Faster\_EMA module. By utilizing pointwise convolution and efficient feature aggregation mechanisms, the model reduces the number of parameters and computational complexity. The SPPF\_CPCA multi-scale structure is designed by introducing the CPCA attention mechanism. By dynamically allocating attention weights, detection accuracy is significantly improved. Additionally, by incorporating multi-scale depthwise separable convolution modules, the model enhances performance in complex environments and under target occlusion conditions. The BiFPN structure is introduced in the feature fusion part, enhancing cross-scale connections and reducing the overall computational burden of the network. Experimental results show that the  PEC-YOLO improves accuracy in power field detection tasks and decreases the overall model size, making it easier to deploy on edge devices. 

\backmatter

\bmhead{Acknowledgements}

The research has been supported by the National Natural Science Foundation of China(Project No.62373144) and the Natural Science Foundation of Shenzhen City(Ref.JCYJ20210324101215039).

\section*{Declarations}
\textbf{Conflict of interest} None

\nocite{*} 
\bibliography{sn-bibliography}

\end{document}